\documentclass[10pt,letterpaper]{article}

\usepackage{ccn}
\usepackage{pslatex}
\usepackage{apacite}

\usepackage{amsmath}
\usepackage{amssymb}
\usepackage{caption}
\usepackage{xcolor} 
\definecolor{Prune}{RGB}{99,0,60} 
\definecolor{B1}{RGB}{49,62,72} 
\definecolor{C1}{RGB}{124,135,143}
\definecolor{D1}{RGB}{213,218,223}
\definecolor{A2}{RGB}{198,11,70}
\definecolor{B2}{RGB}{237,20,91}
\definecolor{C2}{RGB}{238,52,35}
\definecolor{D2}{RGB}{243,115,32}
\definecolor{A3}{RGB}{124,42,144}
\definecolor{B3}{RGB}{125,106,175}
\definecolor{C3}{RGB}{198,103,29}
\definecolor{D3}{RGB}{254,188,24}
\definecolor{A4}{RGB}{0,78,125}
\definecolor{B4}{RGB}{14,135,201}
\definecolor{C4}{RGB}{0,148,181}
\definecolor{D4}{RGB}{70,195,210}
\definecolor{A5}{RGB}{0,128,122}
\definecolor{B5}{RGB}{64,183,105}
\definecolor{C5}{RGB}{140,198,62}
\definecolor{D5}{RGB}{213,223,61}
\usepackage{mdframed}
\usepackage{multirow} 
\usepackage{multicol} 
\usepackage{tikz}
\usepackage{graphicx}
\usepackage[absolute]{textpos} 
\usepackage{colortbl}
\usepackage{array}
\usepackage{geometry}
\usepackage{titlesec}
\usepackage{hyperref}

\usepackage{epstopdf}
\usepackage{epsfig}
\usepackage{rotating}
\usepackage{verbatim}
\usepackage{longtable}
\usepackage{lscape}
\usepackage{subfig}
\usepackage{float}
\usepackage{url}
\usepackage[innercaption]{sidecap}
\usepackage{pdfpages}
\usepackage{xspace}
\usepackage{booktabs} 
\usepackage[authoryear, semicolon, sort&compress]{natbib}
\usepackage{setspace}
\usepackage{nameref}

\title{Probing Brain Context-Sensitivity with Masked-Attention Generation}
 
\author{{\large \bf Alexandre Pasquiou}\\ 
(alexandre.pasquiou@inria.fr) \\
UNICOG, Cognitive Neuroimaging Unit, INSERM, CEA, Neurospin\\
Gif-sur-Yvette, France \\
MIND, INRIA, CEA, Neurospin \\
Gif-sur-Yvette, France
  \AND {\large \bf Yair Lakretz} \\
UNICOG, Cognitive Neuroimaging Unit, INSERM, CEA, Neurospin\\
Gif-sur-Yvette, France
  \AND {\large \bf Bertrand Thirion} \\
MIND, INRIA, CEA, Neurospin \\
Gif-sur-Yvette, France
  \AND {\large \bf Christophe Pallier} \\
UNICOG, Cognitive Neuroimaging Unit, INSERM, CEA, Neurospin\\
Gif-sur-Yvette, France
}

\begin{document}

\maketitle

\clearpage

\section{Abstract} 
{
\bf
Two fundamental questions in neurolinguistics concerns the brain regions that integrate information beyond the lexical level, and the size of their window of integration. 
To address these questions we introduce a new approach named \emph{masked-attention generation}. It uses GPT-2 transformers to generate word embeddings that capture a fixed amount of contextual information.
We then tested whether these embeddings could predict fMRI brain activity in humans listening to naturalistic text. 
The results showed that most of the cortex within the language network is sensitive to contextual information, and that the right hemisphere is more sensitive to longer contexts than the left.
Masked-attention generation supports previous analyses of context-sensitivity in the brain, and complements them by quantifying the window size of context integration per voxel.
}
\begin{quote}
\small
\textbf{Keywords:} 
fmri;transformers; context;brain;encoding
\end{quote}

\section{Introduction}

Following the works of \cite{Bemis2011SimpleCA,Bemis2013-yz}, a few studies have tried to leverage computational models to identify the neural bases of compositionality and quantify brain regions' sensitivity to increasing sizes of context.
Some of them, using ecological paradigms, have found a hierarchy of brain regions that are sensitive to different types of contextual information and different temporal receptive fields \citep[e.g.,][]{wehbe2014aligning, jainincorporating2018, toneva2022combining}.
A notable investigation \citep{jainincorporating2018} used pre-trained LSTM \citep{hochreiterlong1997} models to study context integration. 
They varied the amount of context used to generate word embeddings, and obtained maps indicating brain regions' sensitivity to different sizes of context.
In this work, we study context-sensitivity using the attention mechanisms of GPT-2 which better integrate context than LSTMs.

\section{Methods}

\textbf{fMRI Brain data.}
The brain data consisted of the functional Magnetic Resonance Imaging (fMRI) scans from the English participants of \emph{The Little Prince} fMRI Corpus \citep{LPPdatapaper}\footnote{Available from \url{https://openneuro.org/datasets/ds003643/versions/1.0.2}}.

\begin{figure}[!ht]
    \centering
    \includegraphics[scale=0.75]{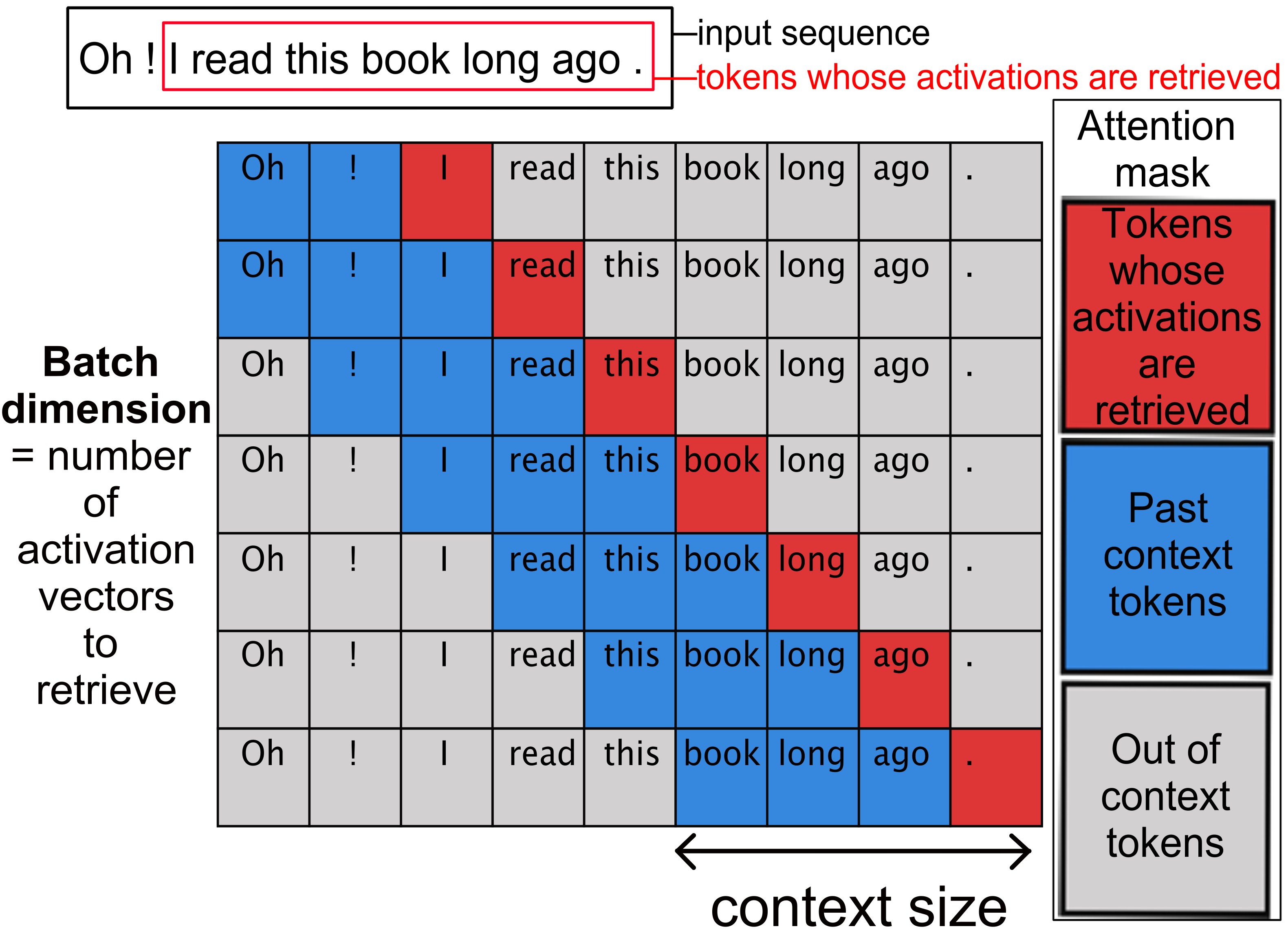}
    \caption[Controlling for tokens' interaction using attention masks.]{
    \textbf{Controlling tokens' interaction using attention masks.}
    Examples of (input sequence, attention mask) pairs to retrieve the embedding of each word of the target sentence (framed in red above).
    An input sequence is represented by a row, the target token is colored in red, tokens in the attention mask are blue or red (context size = 4), and out-of-context tokens are grey.
    }
    \label{c:chap8:fig2}
\end{figure}

\textbf{Modelling Context-limited Features with GPT-2 using attention masks.}
Contextual information was controlled by playing with the attention mechanisms of the GPT-2 \citep{radfordgpt}\footnote{Available from \url{https://huggingface.co/gpt2}} transformer.
The method involves providing the model with an input sequence and attention mask pair for each word in the text, and retrieving the target word's embedding for each pair. 
An example is given in Fig.~\ref{c:chap8:fig2} for a context-window size of 4.

\begin{figure*}[!ht]
    \centering
    \includegraphics[width=\textwidth]{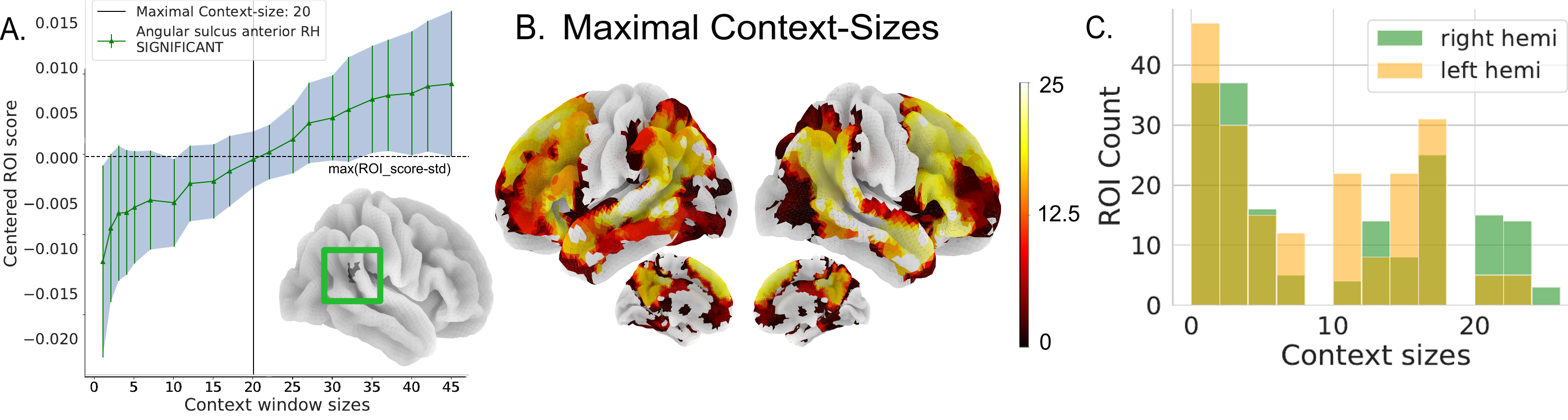}
    \caption[Assessing the maximal context window size over which information is integrated.]{
    \textbf{Assessing the maximal context window size over which information is integrated.}
    A) Determination of the maximal context-size for each parcel of the Difumo atlas. 
    The maximal context-size is defined as the last context-size whose ROI-score is inferior to the maximal averaged centered ROI-score minus its standard deviation.
    B) Surface projection of Difumo's parcels maximal context-size in context-sensitive brain regions.
    C) Histograms representing the maximal context sizes distribution across context-sensitive ROIs, in the left hemisphere (orange), and the right hemisphere (green).
    }
    \label{c:chap8:fig4}
\end{figure*}

The attention mask removed interactions with words outside the window, while preserving interactions within the context window (see Fig.~\ref{c:chap8:fig2}). 
The mask was a binary vector containing 0 except for the target word and the previous n-1 words, where it equaled 1. 
It preserved the positional encoding of words in the sentence and the right use of the special tokens, while using complete sentences.
The attention mask is the same for all the tokens in the input sequence, modulo the incrementality.
Otherwise, information could propagate outside the context window because of model's depth.

\textbf{Encoding models.} The same encoding approach as \cite{pasquiouicml2022,pasquiou2023informationrestricted} was used.
For each context-window size (21 values sampled between 1 and 45 tokens), the embeddings from layer 9 of the 12-layer model (dim=768) were used to fit each subject's brain data (N=51).
Then, we examined the impact of the context-window size on the models' predictive performance ($R$ scores).
The motivation behind this approach is the following.
If the model needs short-range information to build the embedding of a word, then the embedding won't be affected when using a small context size. 
However, if the model needs long-range information, the embedding will be `damaged' when using a small context size.
Thus, increasing context size won't improve $R$ scores in the brain regions well-fitted by features using short-range information.
However, brain regions well-fitted by features using long-range information will benefit from increasing context size.

\textbf{DIFUMO atlas.}
We computed the median $R$ score across voxels constituting 90\% of the non-zero loadings of each parcel of the Difumo atlas \citep{DADIDIFUMO} (referred to as \emph{ROI-score}).

\textbf{Assessing brain regions sensitivity to context.}
For each participant and ROI, we fitted a Linear Regression on the (context\_size, ROI-score) points to get the slope of increase of the ROI-score as a function of context-size. 
Brain regions' context-sensitivity was estimated with a t-test on the slopes of increase across subjects, with a FDR correction of 0.01 to account for multiple comparisons.

\textbf{Quantifying the window-size over which context is integrated.}
For each context-sensitive parcel of the atlas, we estimated its \textit{maximal context-size}, i.e. the last context-window size over which the ROI-score is less than one standard deviation away from its maximal value (Fig.~\ref{c:chap8:fig4}A).
Maximal context-sizes are reported in Fig.~\ref{c:chap8:fig4}B.

\section{Results \& Discussion}
First, \textbf{most of the language related brain regions are context-sensitive} (Fig.~\ref{c:chap8:fig4}B).
This network of context-sensitive brain regions is bilateral and mostly symmetrical.
Notes that low-level regions such as the auditory, motor and visual cortices are not context-sensitive.
These findings support the ones from \cite{jainincorporating2018}.

Additionally, we found that \textbf{the right hemisphere shows sensitivity to longer contexts than the left} (Fig.~\ref{c:chap8:fig4}C).
The brain regions integrating longer-context revolves around the Temporo-Parietal Junction, Superior frontal regions and medial regions.
This observation is consistent with other brain imaging studies that have supported the role of the right hemisphere in higher-level language tasks (see \citet{jungbeemanbilateral2005,beeman2013}).
Overall, our results show that modifications of language models' architecture (e.g., unit ablation), or internal operations (e.g., modification of the attention mechanisms) can be used to probe precise linguistic processes.

\clearpage

\section{Acknowledgments}
This project/research has received funding from the American National Science Foundation under Grant Number 1607441 (USA), the French National Research Agency (ANR) under grant ANR-14-CERA-0001, the European Union’s Horizon 2020 Framework Programme for Research and Innovation under the Specific Grant Agreement No. 945539 (Human Brain Project SGA3), and the KARAIB AI chair (ANR-20-CHIA-0025-01).

\bibliographystyle{apacite}

\setlength{\bibleftmargin}{.125in}
\setlength{\bibindent}{-\bibleftmargin}

\bibliography{main}

\end{document}